\documentclass{ifacconf}

\usepackage{graphicx}
\usepackage{xcolor}
\usepackage{subcaption}
\usepackage{booktabs}
\usepackage{tabularx}
\usepackage{array}
\usepackage{xurl}
\usepackage{amsmath,amsfonts,amssymb}
\usepackage{tikz}
\usepackage[font=small]{caption}
\usepackage{pgfplots}
\usepgfplotslibrary{groupplots}
\usepackage{natbib}
\usepackage{enumitem}
\usepackage{multirow}

\makeatletter
\long\def\@makefigurecaption#1#2{%
  \vskip 3pt%
  {\footnotesize\par\hangindent2em\@fignumfont{#1}. #2\par}%
  \vskip 2pt%
}
\makeatother

\newcommand{\SO}{SO(3)}

\newcommand{\SE}{SE(3)}
\newcommand{\se}{se(3)}

\newcommand{\fg}{f_{_G}}
\newcommand{\eg}{e_{_G}}
\newcommand{\ev}{e_V}
\newcommand{\Ad}{\text{Ad}}
\newcommand{\tr}{\text{tr}}

\begin{document}
\begin{frontmatter}

\title{\vspace{-10pt} Geometric Formulation of Unified Force-Impedance Control on $SE(3)$ for Robotic Manipulators} 
% Title, preferably not more than 10 words.

\thanks[footnoteinfo]{Joohwan Seo, Soomi Lee, Arvind Kruthiventy, and Roberto Horowitz were partially funded by the Hong Kong Center for Construction Robotics Limited. Jongeun Choi was supported by the National Research Foundation of Korea(NRF) grant funded by the Korea government(MSIT) (No.RS-2024-00344732).}

\author[First]{Joohwan Seo} 
\author[First]{Nikhil Potu Surya Prakash} 
\author[First]{Soomi Lee}
\author[First]{Arvind Kruthiventy}
\author[First]{Megan Teng}
\author[Second]{Jongeun Choi}
\author[First]{Roberto Horowitz}

\address[First]{University of California, Berkeley, USA}
\address[Second]{Yonsei University, Seoul, Republic of Korea
\\ e-mail: \{joohwan\_seo, nikhilps, soomi\_lee, arvindkruthiventy, meganteng, horowitz\}@berkeley.edu, 
jongeunchoi@yonsei.ac.kr
}

\begin{abstract}                % Abstract of 50--100 words
In this paper, we present a geometric unified force–impedance control (GUFIC) framework on the $\SE$ manifold that enables force tracking while guaranteeing passivity. 
Building upon unified force–impedance control (UFIC) and geometric impedance control (GIC), 
GUFIC incorporates the $\SE$ manifold structure through a differential–geometric formulation and augments energy tanks for both force-tracking and impedance control to ensure closed-loop passivity. 
The proposed framework resolves the implementation difficulty of UFIC by introducing velocity and force fields, which enable causal updates of desired motion and force. 
Defined entirely on $\SE$, GUFIC inherits the $\SE$ invariance and equivariance properties of GIC, improving generalization and sample efficiency when integrated with learning-based policies. 
The proposed control law is validated in a
simulation environment under scenarios requiring tracking an
$\SE$ trajectory, incorporating both position and orientation, while exerting a force on a surface. 
The implementation is available at \url{https://github.com/Joohwan-Seo/GUFIC_mujoco}.
\end{abstract}


\end{frontmatter}
%===============================================================================

\section{Introduction}\label{Sec:Intro}
After its introduction by \cite{hogan1985impedance}, impedance control has become a fundamental framework for robotic manipulation involving interaction with uncertain environments. Frequently paired with the operational space formulation \citep{khatib1987unified}, it remains a standard method for end-effector control. Because of its ability to ensure safe physical interaction, impedance/admittance control is often adopted as the low-level controller in recent learning-based manipulation policies \citep{bogdanovic2020learning}.

% Recent advancements in deep-learning approaches showed impressive results in performing real-life tasks. In particular, imitation learning-based policies, such as behavior transformer \cite{shafiullah2022behavior}, diffusion policy \cite{chi2023diffusion}, action-chunking transformer \cite{zhao2023learning}, and their variants, have demonstrated success by producing desired end-effector trajectories from vision inputs. 

Recent advances in deep learning have achieved impressive real-world performance. In particular, imitation learning–based policies such as Behavior Transformer \citep{shafiullah2022behavior}, Diffusion Policy \citep{chi2023diffusion}, and Action-Chunking Transformer \citep{zhao2023learning} have demonstrated successful robotic task execution by producing desired end-effector trajectories from visual inputs.

However, policies that output only desired poses remain insufficient for high-precision, contact-rich tasks. 
To address this limitation, recent works predict impedance/admittance gains \citep{zhou2024admittance, kamijo2024learning, seo2025equicontact} or direct force profiles \citep{wu2024tacdiffusion}. 
While modifying gains or applying direct force alongside the desired pose can enhance performance in contact-rich scenarios, these approaches introduce challenges in maintaining stability.
From a control-theoretic perspective, the stability guarantees of static-gain impedance/admittance controllers no longer hold when the gains vary during execution \citep{kronander2016stability}.

The stability analysis of a robot interacting with the environment is closely tied to the concept of passivity. According to the passivity theorem, if a robot's closed-loop dynamics are designed to be passive, the overall robot-environment interaction remains stable, since the environment itself is strictly passive \citep{li2002passive,li2001PVFCP1,li2001PVFCP2}. \cite{li1997SmartEscersiceMachinesP1,li1997SmartEscersiceMachinesP2} illustrate the use of closed-loop passivity in the design and implementation of learning controllers for self-optimizing exercise machines, where safe human-machine interaction is indispensable. Other passivity-based control designs can be found in \cite{ferraguti2015energy, rashad2019energy, michel2020passivity}.

% In \cite{haddadin2024unified}, the unified force-impedance control (UFIC) was proposed as a means for a robot manipulator to maintain contact with the environment while executing a task using impedance control and exerting a desired force. Using energy tank augmentation, the UFIC control ensures the passivity of the closed-loop system. 
The unified force–impedance control (UFIC) \citep{haddadin2024unified} was introduced to enable a robot manipulator to maintain contact with the environment while exerting a desired force through impedance control. By augmenting the control law with energy tanks, UFIC guarantees the passivity of the closed-loop system.
% However, the UFIC does not consider the $\SE$ manifold structure inherent in the manipulator's end effector pose description and simply treats pose misalignment errors as Cartesian vectors. Another drawback in \cite{haddadin2024unified} is that, in order to establish the passivity of the impedance control term using tank augmentation, a modified desired velocity must be defined, and then it must be integrated to generate a new trajectory. This complicates the update of the next step's desired velocity and could lead to a causality breakdown in the control process. This is further discussed in Section \ref{sec:VFfield}.
However, the UFIC did not consider $\SE$ manifold structure of the end-effector, treating pose errors as simple Cartesian vectors.
Moreover, to preserve passivity via tank augmentation, UFIC requires defining and integrating a modified desired velocity to generate a new trajectory, which complicates online updates and may induce causality issues (see Section~\ref{sec:VFfield}).
% \ref{sec:tankimpedance}.

% Our previous work on geometric impedance control (GIC) \cite{seo2023geometric, seo2024comparison} established a unified framework for controlling the end-effector's position and orientation using differential geometry. 
Our prior work on geometric impedance control (GIC) \citep{seo2023geometric, seo2024comparison} established a differential-geometric framework for unified position–orientation control on $\SE$.
% A significant advantage of the GIC framework is that, as shown in \cite{seo2023contact, seo2025equicontact}, considering the $\SE$ manifold structure in the control structure leads to $\SE$ invariance and equivariance in the learned policy, which greatly enhances spatial generalizability. 
As shown by \cite{seo2023contact, seo2025equicontact}, leveraging the $\SE$ manifold structure yields $\SE$ invariant and equivariant control policies with strong spatial generalization. 
Recent studies have further confirmed that $\SE$ equivariance improves sample efficiency and robustness to out-of-distribution data in visual manipulation learning \citep{ryu2022equivariant, ryu2023diffusion, huang2024fourier, seo2025se}.

% In this paper, we present a geometric formulation of the unified force-impedance control scheme on the $\SE$ manifold for a robotic manipulator, namely a geometric unified force-impedance control (GUFIC). 
In this paper, we present a geometric formulation of unified force–impedance control on the $\SE$ manifold, named geometric unified force–impedance control (GUFIC).
The main contributions are summarized as follows:
\begin{enumerate}[leftmargin=*]
    \item Unlike UFIC \citep{haddadin2024unified}, which treats translational and rotational errors as independent vector-space elements, GUFIC consistently incorporates both within the $\SE$ Lie group structure.
    % we fully incorporate the manifold structure of $\SE$, and handled  translation and orientation errors consistently within the $\SE$ Lie group structure.
    \item We introduce a time-dependent velocity field to encode tasks and generate trajectories while preserving causality, resolving the implementation difficulties of \cite{haddadin2024unified}
    \item Following the formulation by \cite{seo2023contact}, the resulting control law is $\SE$ equivariant, enhancing spatial generalizability for learning manipulation tasks.
    % Since we follow the formulation of \cite{seo2023contact}, the resulting control law is $\SE$-equivariant, thus providing further advantages in learning transferability and sample efficiency for learning manipulation tasks.
    \item The passivity of the control law ensures contact stability and safety in force-based manipulation tasks.
    % From the perspective of the learning manipulation tasks incorporating forces, the passive behavior of the control formulation will further provide advantages such as contact stability.
\end{enumerate}
\textbf{Problem Formulation}\;
Let \(g(t)\in SE(3)\) denote the end-effector pose and \(V^b(t)\in\mathbb{R}^6\) its body-frame twist. 
We aim to design a controller that tracks a desired trajectory while regulating
contact force and preserving closed-loop passivity.
We assume that the following desired trajectory $g_d(t)$ and force field $F_d(t,g)$ are known a priori:
\begin{equation}\label{eq:desired_trajectory_force_field}
g_d(t):\mathbb{R}_{\ge0}\to SE(3),\quad
F_d(t,g):\mathbb{R}_{\ge0}\times SE(3)\to\mathbb{R}^6 .
\end{equation}
The objective is to construct a force-impedance controller that is compatible with the $\SE$ manifold structure, causally implementable, and passive during contact.

% The remainder of this paper is organized as follows: Section~\ref{Sec:Prelim} reviews the necessary background on Lie groups, manipulator dynamics, and geometric impedance control (GIC); Section~\ref{Sec:GUFIC} presents the GUFIC framework and its passivity proof; Section~\ref{Sec:Simulation} reports simulation results; and Section~\ref{Sec:Conclusion} concludes the paper.

%
% \section{Problem Formulation}\label{Sec:Problem}
% \input{Sections/sec_problem}
%
\section{Preliminaries}\label{Sec:Prelim}
\subsection{$SE(3)$ Notation and Body-frame Dynamics}
The end-effector configuration is represented by 
$g=(p,R)\in\SE$, where $p\in\mathbb{R}^3$ and $R\in\SO$. 
The corresponding body-frame twist is denoted by
\begin{equation}
    V^b = 
    \begin{bmatrix} v^b \\ \omega^b \end{bmatrix}
    = J_b(q)\dot{q},
\end{equation}
where $J_b(q)$ is the body Jacobian. We use $\widehat{(\cdot)}$ and 
$(\cdot)^\vee$ for the standard hat and vee maps.

The $\SE$ operational-space dynamics of the end effector, expressed in 
the body frame, are written as
\begin{equation}\label{eq:robot_dynamics_eef}
    \tilde{M}(q)\dot{V}^b + \tilde{C}(q,\dot{q})V^b + \tilde{G}(q) = F + F_e,
\end{equation}
where $F$ is the commanded body-frame wrench and $F_e$ is the external 
wrench exerted by the environment. 
For compactness, we omit the 
dependencies on $q$ and $\dot q$ unless needed, e.g., 
$\tilde{M}=\tilde{M}(q)$ and $\tilde{C}=\tilde{C}(q,\dot q)$. Due to the page limit, we will refer to the detailed definitions of $\tilde{M}$, $\tilde{C}$, and $\tilde{G}$ in \cite{seo2023geometric, seo2024comparison}. 

\subsection{Geometric Impedance Control}
In our previous work \citep{seo2023geometric, seo2024comparison}, geometric impedance 
control (GIC) was proposed for the tracking control of the end-effector pose on $\SE$. 
The GIC law is given by
\begin{equation}\begin{aligned}\label{eq:GIC_law}
    F_i = \tilde{M} \dot{V}_d^* + \tilde{C} V_d^* + \tilde{G} 
    - \fg(g,g_d) - K_d \ev,
\end{aligned}\end{equation}
where $g_d=(p_d,R_d)$ is the desired configuration, 
$K_d\in\mathcal{S}^6_{++}$, and $\ev=V^b-V_d^*$. The translated desired 
body-frame velocity is defined by
\begin{equation}
    V_d^* = \Ad_{g_{ed}}V_d^b,\qquad g_{ed}=g^{-1}g_d,
\end{equation}
with
\begin{equation}
    \Ad_g = 
    \begin{bmatrix}
        R & \hat{p}R \\ 0 & R
    \end{bmatrix}.
\end{equation}
The elastic wrench $\fg$ is defined as
\begin{equation}\begin{aligned} \label{eq:geometric_impedance}
    \fg(g,g_d) \!=\! \begin{bmatrix}
    R^T R_d K_p R_d^T (p \!-\! p_d)\\
    (K_R R_d^T R \!-\! R^T R_d K_R)^\vee
    \end{bmatrix},
\end{aligned}\end{equation} 
where $K_p,K_R\in\mathcal{S}^3_{++}$. The associated potential and 
kinetic energy functions are
\begin{align}\label{eq:potential_and_kinetic_energy}
    &P(g,g_d) \!=\! \tr(K_R(I \!-\! R_d^TR))
              + \tfrac{1}{2}(p \!-\! p_d)^T R_d K_p R_d^T(p \!-\! p_d), 
              \nonumber\\
    &K(\ev) \!=\! \tfrac{1}{2} \ev^T \tilde{M} \ev.
\end{align}

The GIC law is formulated based on the following assumption.
\begin{assum}[\cite{seo2023geometric}]
    \label{assum:1}
    The body Jacobian $J_b$ is full-rank in the operating region, and 
    both the actual and desired end-effector trajectories remain in the 
    reachable set. The desired trajectory is continuously differentiable.
\end{assum}
\section{Geometric Unified Force Impedance Control}\label{Sec:GUFIC}
% In this section, we will present the detailed walkthrough to re-formulate the original UFIC to fully incorporate the $\SE$ manifold structure of the end-effector. 
In this section, we present a detailed walkthrough to reformulate the original UFIC framework to fully incorporate the $\SE$ manifold structure of the end effector.
% \subsection{Naive Force Tracking Control Law}
% The first step is to augment the geometric impedance control $F_i$ in \eqref{eq:GIC_law} with the force tracking control $F_f$ so that the total input control wrench is 
% \begin{equation} \label{eq:naive}
%     F = F_i + F_f,
% \end{equation}
\subsection{Naive Force Tracking Control Law}
We begin by augmenting the geometric impedance control $F_i$ in \eqref{eq:GIC_law} with a force-tracking term $F_f$, such that the total control wrench is
\begin{equation}\label{eq:naive}
    F = F_i + F_f,
\end{equation}
where $T_f = J_b^T F_f$. The force controller is defined in the PID form as
\begin{equation}\label{eq:force_control}
    \begin{split}
        F_f =& -k_p e_f
               - k_d \frac{d}{dt}e_f  - k_i \!\int \!e_f(\tau)\, d\tau
               + F_d,
    \end{split}
\end{equation}
where $e_f(t) = -\bar{F}_e(t) - F_d(t)$ is wrench error and $\bar{F}_e = \bar{F}_e(t)$ denotes the force/torque sensor measurement, consistent with \cite{haddadin2024unified} and $\bar{F}_e$ is used interchangeably with $F_e$, the actual external wrench exerted by the environment. In addition, $F_d$ is the desired force field introduced in \eqref{eq:desired_trajectory_force_field}.  
However, as will be shown in the following, the naive force–impedance control law \eqref{eq:naive} does not guarantee passivity.
% As in \cite{haddadin2024unified},  we use $\bar{F}_e = \bar{F}_e(t)$ for the force/torque sensor output, without discerning it from $F_e$ the actual external wrench that the environment exerts on the robot $F_e$. In addition, $F_d$ is the desired force field introduced in \eqref{eq:desired_force_field}.
% However, as will be shown in the following, the naive force-impedance control law \eqref{eq:naive} is not passive.

\subsubsection{Passivity Analysis}
To analyze the passivity of the control system, we first recall its definition.
\begin{defn} \label{def:passivity}
    The control system is passive with respect to the pair $(V^b, F_e)$, or the supply rate $(V^b)^T\! F_e$, when the following condition is satisfied:
    \begin{equation}
        \dot{S} \leq (V^b)^T F_e,
    \end{equation}
    where $S \in \mathbb{R}_{\geq 0}$ is a positive definite storage function.
\end{defn}
The error dynamics with the naive geometric force-impedance control law can be written as
\begin{equation}
    \tilde{M} \dot{e}_V + \tilde{C} e_V + K_d e_V + \fg - F_f - F_e = 0.
\end{equation}

We define the storage function as the sum of kinetic $K$ and potential $P$ energies on the $\SE$ group defined earlier \eqref{eq:potential_and_kinetic_energy}:
% Following the formulation from \cite{seo2023geometric}, we will use the summation of the potential energy function in $\SE$ and kinetic energy function as the storage function, i.e.,
\begin{equation}
    \begin{split}
    &S(t,q,\dot{q}) = K(t,q,\dot{q}) + P(t,q).
    \end{split}
\end{equation}
Using the fact that $\dot{P} = \fg ^T \ev$ from \cite{seo2023geometric, seo2024comparison}, one can further show that
\begin{align}
    &\dfrac{dS}{dt} = \ev^T \tilde{M} \dot{e}_V + \dfrac{1}{2} \ev^T \dot{\tilde{M}}\ev + \fg^T \ev \nonumber \\
    &= \ev^T (- \tilde{C} \ev - K_d \ev - \fg + F_f + F_e + \fg) + \tfrac{1}{2} \ev^T \dot{\tilde{M}} \ev \nonumber \\
    &= \underset{=0}{\underbrace{\ev^T (\tfrac{1}{2} \dot{\tilde{M}} - \tilde{C}) \ev}} - \underset{\geq 0}{\underbrace{\ev^T K_d \ev}} + \ev^T F_f + \ev^T F_e \nonumber \\
    & \leq  (V^b)^T F_e + (V^b)^T F_f - (V_d^*)^T(F_f + F_e).
\end{align}
Since the signs of the terms $(V^b)^T F_f$ and $(V_d^*)^T (F_f + F_e)$ are not determined, the passivity of the control system with naive force tracking control law cannot be guaranteed.

\subsection{Passive Control Law Formulation}
To ensure closed-loop passivity, we incorporate energy storage via tank augmentation for both the force-tracking and impedance controllers, following \cite{haddadin2024unified}.

\subsubsection{Tank Augmentation for Force-tracking Controller}
First, for the port $(V^b, F_f)$, the energy tank with respect to the force control $T_f$ is first defined as
\begin{equation}
    T_f = \dfrac{1}{2} x_{tf}^2, \quad x_{tf} \neq 0,
\end{equation}
with the force control tank state variable $x_{tf}$ and its dynamics
\begin{equation} \label{eq:force_tank_state_dyanmics}
    \dot{x}_{tf} = -\dfrac{\beta_f}{x_{tf}} \gamma_f (V^b)^T F_f + \dfrac{\alpha_f}{x_{tf}} (\gamma_f -1) (V^b)^T F_f,
\end{equation}
where
{\small
\begin{align*}
    \gamma_f \!&=\! \left\{\begin{array}{lc}
       1  &  \text{if}\;  (V^b)^T F_f < 0\\
       0  & \text{otherwise}
    \end{array}\right., \;
    \beta_f \!=\! \left\{\begin{array}{lc}
       1  &  \text{if}\; T_f \leq T^{u,f}\\
       0  & \text{otherwise}
    \end{array}\right. \nonumber \\
    \alpha_f \!&=\! \left\{\begin{array}{lc}
        1 & \text{if} \; T_f \geq T_{l,f} + \delta_{T,f} \\
        \tfrac{1}{2}\! \left(1 \!-\! \cos{\left(\tfrac{T_f - T_{l,f}}{\delta_{T,f}} \pi\right) }\right) & \text{if} \; T_{l,f} + \delta_{T,f} \geq T_f \geq T_{l,f} \\
        0 & \text{otherwise}.
    \end{array}
    \right. \nonumber
\end{align*}
}
% where $T^{u,f}$ and $T_{l,f}$ denote the upper limit and lower limit of the energy tank for force tracking control, respectively, and $\delta_{T,f}$ is a margin towards the lower limit of the energy tank to enable smooth switching behavior.
Here, $T^{u,f}$ and $T_{l,f}$ denote the upper and lower limits of the energy tank, and $\delta_{T,f}$ provides a smooth transition margin near the lower bound.  
% The variables $\gamma_f$, $\beta_f$, and $\alpha_f$ indicate whether the controller injects or dissipates energy, thereby preserving passivity.  
The purpose of $\gamma_f$ is an indicator of whether the force tracking law $F_f$ is in the passivity-violating direction, and $\beta_f$ is to prevent the overflow of the energy tank.
The resulting modified force command is
\begin{equation} \label{eq:force_tracking_modified}
    F_f' = (\gamma_f + \alpha_f (1 - \gamma_f)) F_f,
\end{equation}
which scales the force output according to the available tank energy, ensuring that the port $(V^b, F_f')$ remains passive.

\subsubsection{Tank Augmentation for Impedance Controller}
\label{sec:tankimpedance}A similar mechanism regulates the energy exchange of the motion-tracking port $(V_d^*, -(F_f' + F_e))$.  
The impedance control tank is defined as
\begin{equation}
    T_i = \tfrac{1}{2} x_{ti}^2, \quad x_{ti} \neq 0,
\end{equation}
with impedance tank state $x_{ti}$ and its dynamics
\begin{equation}\label{eq:impedance_tank_state_dynamics}
    \begin{split}
        \dot{x}_{ti} &=
        \tfrac{\beta_i}{x_{ti}}\!\left[\gamma_i(V_d^*)^{\!T}(F_f' + F_e)
        + (e_V')^{\!T}K_d e_V'\right] \\
        &\quad + \tfrac{\alpha_i}{x_{ti}}(1 - \gamma_i)
        (V_d^*)^{\!T}(F_f' + F_e),
    \end{split}
\end{equation}
where
\begin{equation*}
    e_V' = V^b - (\gamma_i + \alpha_i(1 - \gamma_i))V_d^*
           = V^b - V_d',
\end{equation*}
and
\begin{equation}\label{eq:desired_velocity_modification}
    V_d' = (\gamma_i + \alpha_i(1 - \gamma_i))V_d^*
\end{equation}
is the modified desired velocity. $\gamma_i$, $\beta_i$ and $\alpha_i$ are defined respectively as
{ \small
\begin{equation*}
    \begin{split}
    \gamma_i \!&=\! \left\{\begin{array}{lc}
       1  &  \text{if}\;  (V_d^*)^T (F_f' + F_e) \!>\! 0\\
       0  & \text{otherwise}
    \end{array}\right., \;
    \beta_i \!=\! \left\{\begin{array}{lc}
       1  &  \text{if}\; T_i \!\leq\! T^{u,i}\\
       0  & \text{otherwise}
    \end{array}\right.\\
    \alpha_i \!&=\! \left\{\begin{array}{lc}
        1 & \text{if} \; T_i \geq T_{l,i} + \delta_{T,i} \\
        \tfrac{1}{2}\! \left(1 \!-\! \cos{\left(\tfrac{T_i - T_{l,i}}{\delta_{T,i}} \pi\right) }\right) & \text{if} \; T_{l,i} + \delta_{T,i} \geq T_i \geq T_{l,i} \\
        0 & \text{otherwise},
    \end{array}
    \right.
    \end{split}
\end{equation*}
}
where the variables $T^{u,i}$, $T_{l,i}$, and $\delta_{T,i}$ can be understood analogously to the ones for the force tracking counterparts.

While this augmentation guarantees passivity, it introduces an implementation issue: the modified velocity $(V_d^*)'$ must be integrated to update the entire reference trajectory, which may require recursive iteration. 
This motivates the introduction of a time- and state-dependent velocity-field formulation, presented in the following subsection.

\subsection{Velocity Field and Force Field Formulation}
\label{sec:VFfield}
\subsubsection{Velocity Field}
% One subtle challenge arises when implementing the modified desired velocity $V_d'$: updating the corresponding signals $\dot{V}_d'$ and $g_d'$ accordingly. While the original work \cite{haddadin2024unified} suggested integrating and differentiating the modified velocity signal appropriately, naive implementations can easily lead to causality problems, requiring careful engineering to ensure precise updates. Specifically, integrating the modified velocity at the current time step yields an updated trajectory, yet the next-step desired velocity cannot be directly computed in a causal manner. 
A key implementation challenge arises when updating the modified desired velocity $V_d'$ and its corresponding signals $\dot{V}_d'$ and $g_d'$. 
While the original work \citep{haddadin2024unified} suggested integrating and differentiating the modified velocity appropriately, naive implementation procedures may violate causality. To elaborate, integrating $V_d'(k)$ at the current time $k$ yields an updated trajectory $g_d'(k+1)$, but the subsequent computation of $V_d^b(k+1)$ depends on $g_d'(k+1)$, which may lead to a recursive dependency that complicates causal implementation.
To address this, we reinterpret the desired motion as a \emph{velocity field} \citep{li2002passive}, defined over both time and configuration. 

Formally, the velocity field $V_d^*(t,g)$ is defined as a mapping from the current time $t$ and the current pose $g \in \SE$ to vectors on the Lie algebra $\se$, ensuring $g(t)\hat{V}_d^*(t,g)\in T_g\SE$. Thus, our proposed velocity field is expressed as $\hat{V}_d^*: \mathbb{R}_{\geq0}\times\SE \to \se$, explicitly dependent on both time and pose rather than solely time.

Let $g_d(t)=(p_d(t),R_d(t))$ denote the nominal desired trajectory,
with desired body-frame velocity $\hat{V}_d^b(t)=g_d^{-1}\dot{g}_d$.
We drop the time dependency when it is clear. Following \cite{li2002passive},
the time-varying velocity field is defined as
% Let us first denote the original desired time-trajectory from \eqref{eq:desired_trajectory_force_field} as $g_d(t) = (p_d(t), R_d(t))$ and their corresponding desired body-frame velocity as $\hat{V}_d^b(t)$. We will also drop the time dependency to avoid clutter, e.g., $g_d = g_d(t)$. 
% Following the formulation of \cite{li2002passive}, the time-varying velocity field $\hat{V}_d(t,g)$ from the desired trajectory $g_d$ and $\hat{V}_d^b = g_d^{-1} \dot{g}_d$ is given by
\begin{equation} \label{eq:velocity_field}
    \hat{V}_d(t,g) = g_{ed} \hat{V}_d^b(t) g_{ed}^{-1} + \zeta \nabla_1 \Psi(g,g_d),
\end{equation}
where $g_{ed} = g^{-1}g_d$, and $\nabla_1$ denotes the gradient to its first argument. 
$\Psi(g,g_d)$ is an error function given by
\begin{equation*} 
    \begin{split}
        \Psi(g,g_d) &= \frac{1}{2}\|I_{4} - g_d^{-1}g\|_F^2
        = \tr(I - R_d^TR) + \tfrac{1}{2}\|p - p_d\|_2^2,
    \end{split}
\end{equation*}
that can serve as an error function on $\SE$, proposed in \cite{seo2023geometric}. $\zeta \in \mathbb{R}_{>0}$ is a positive scalar gain. One can also notice that from \cite{murray1994mathematical} (see Chapter 2.4 of it),
\begin{equation*}
    \hat{{V}}_d^* = {g}_{ed}\hat{{V}}_d^b{g}_{ed}^{-1} \iff {V}_d^* = \Ad_{{g}_{ed}}{V}_d^b.
\end{equation*}
Additionally, it is shown in \cite{seo2023geometric} that
\begin{equation}
    \nabla_1\Psi(g,g_d) = \hat{e}_{_G}(g, g_d),
\end{equation}
where $\eg(g,g_d)$ is a geometrically consistent error vector (GCEV) proposed in \cite{seo2023contact, seo2023geometric} given by
\begin{equation}
    e_{_G}(g,g_d) = \begin{bmatrix}
        R^T(p - p_d) \\
        (R_d^T R - R^TR_d)^\vee
    \end{bmatrix},
\end{equation}
see also \cite{bullo1999tracking, prakash2024deep, prakash2024variable} for more results on $\SE$, and \cite{lee2010geometric} for the result on $\SO$. 

Now that we are equipped with the velocity field $V_d(t,g)$, the velocity modification law \eqref{eq:desired_velocity_modification} can be freely applied without harming causality, by replacing $V_d^*$ with $V_d$. In particular, the formula for the modified desired velocity \eqref{eq:desired_velocity_modification} is updated as
\begin{equation}
    V_d' = (\gamma_i + \alpha (1-\gamma_i))V_d
\end{equation}
Using the modified velocity field for the trajectory tracking $\hat{V}_d'(t,g)$, the modified desired configuration $g_d'(t)$ can be obtained by integrating from
\begin{equation} \label{eq:modified_setpoint_dynamics}
    \dot{g}_d' = g_d' (\hat{V}_d^b)',
\end{equation}
where $(V_d^b)' = \Ad_{g_{ed}'^{-1}}V_d'$, with $g_{ed}' = g^{-1}g_d'$.
In discrete form,
\begin{equation*}
    g_d'(k+1) \approxeq g_d'(k) \exp{\left((\hat{V}^b_d)'(k) \Delta t\right)}.
\end{equation*}
The time derivative $\dot V'_d$ is obtained by differentiating the velocity field in \eqref{eq:desired_trajectory_force_field}; the explicit expression is omitted for space and is provided in the released implementation.
% For the full control law derivation, the time derivative of the velocity field $V_d$ needs to be directly calculated. The full calculations for  $V_d$ and  $\dot{V}_d$ are shown in the Appendix.~\ref{Sec:Appendix}.

\subsubsection{Force Field}
Similarly, the desired wrench can be viewed as a force field
$F_d(t,g)\in T_g^*\SE$, where $T_g^*\SE$ denotes the dual space of $T_g\SE$. If a desired wrench is specified at the desired pose
$g_d$, it can be transported to the current body frame using the dual adjoint
map, 
\begin{equation}
    F_d^*=\Ad^*_{g_{ed}^{-1}}F_d = \Ad_{g_{ed}^{-1}}^T F_d \in T_g^* \SE,
\end{equation}
where $F_d^*$ denotes the translated desired body-frame wrench to the current frame, similar to $V_d^*$. In this paper, we consider a simple case: a constant desired normal force in the simulation.

% \subsubsection{Force Field}
% Analogous to the velocity field formulation, the desired force can also be represented as a field defined on the manifold. 
% To achieve maximum generality, we define the desired force as a time- and configuration-dependent mapping, i.e., $F_d = F_d(t,g) \in T_g^*\SE$, where $T_g^*\SE$ denotes the dual space of $T_g\SE$. Based on this formulation, one can use a time-dependent desired force by dropping the dependency on the current configuration $g$, and vice versa. The other consideration is whether to define the desired force on the point $g$ or $g_d$ on the manifold on $\SE$. When the desired force is represented on the dual space of the desired pose $g_d$, $F_d(t) \in T_{g_d}^*\SE$, the appropriate coordinate transformation, i.e., dual Adjoint map $\Ad_{g^{-1}_{ed}}^*$  \citep{seo2023contact}, needs to be applied, so that:
% \begin{equation*}
%     F_d^*(t) = \Ad_{g^{-1}_{ed}}^*F_d(t) = \Ad_{g_{ed}^{-1}}^T F_d(t) \in T_g^*\SE,
% \end{equation*}
% where $F_d^*$ denotes the translated desired body-frame wrench to the current frame, similar to $V_d^*$.

% In this paper, we consider the simple case where $F_d$ is constant over both state and time. 

\subsection{Final Geometric Unified Force-Impedance Controller}
Using the modified setpoint calculated from the velocity field, the modified impedance controller is formulated as follows:
\begin{equation} \label{eq:modified_GIC}
    F_i' = \tilde{M} \dot{V}_d' + \tilde{C} V_d' + \tilde{G} - \fg(g, g_d') - K_d \ev'.
\end{equation}
The associated storage function now reads:
\begin{equation} \label{eq:storage_function}
    S = \dfrac{1}{2} (\ev')^T \tilde{M} \ev' + P(g,g_d').
\end{equation}
Combining the force tracking and impedance control components yields the final GUFIC law:
\begin{equation} \label{eq:passive_GUFIC}
    T = J_b^T F', \;\;\text{where}\; F' = F_f' + F_i',
\end{equation}
where $F_f'$ and $F_i'$ are shown in \eqref{eq:force_tracking_modified} and \eqref{eq:modified_GIC}, respectively. The main theorem of the paper is presented:

\begin{thm}[Passivity of the GUFIC]
    Suppose that assumption~\ref{assum:1} holds true. Consider a robotic manipulator with dynamics \eqref{eq:robot_dynamics_eef} and the GUFIC control law \eqref{eq:passive_GUFIC}. Then, the closed-loop system is passive for the channel $(V^b, F_e)$, with respect to the total storage function $S_{tot}$ given by:
    \begin{equation} \label{eq:total_storage_function}
        S_{tot} = S + T_i + T_f.
    \end{equation}
    Specifically, the following equation holds:
    \begin{equation} \label{eq:time_deriv_storage_passive}
        \dot{S}_{tot} \leq (V^b)^T F_e
    \end{equation}
\end{thm}
The proof is presented in Appendix~\ref{Sec:proof}.

\subsection{$\SE$ Invariance and Equivariance}
The proposed GUFIC framework is also equipped with $\SE$ invariant and equivariant structure. To prove equivariance, two conditions are needed \citep{seo2023contact}: 1) Left invariance, and 2) a control law defined on the body frame.
The terms related to the GIC satisfy the left-invariance condition, e.g., the elastic force $\fg$ is left invariant:
\begin{equation}
    \fg(g,g_d) = \fg(g_lg, g_lg_d), \quad \forall g_l \in \SE,
\end{equation}
see Lemma~1 in \cite{seo2023contact}. The velocity and force field are defined in their own body frames, i.e., left-invariant vector/covector fields. 
Since the GUFIC laws are based on left-invariant vector/covector fields and the elastic control is left-invariant, the resulting GUFIC law is left-invariant. 
Moreover, as the control law is left-invariant and defined on the end-effector body frame, it is $\SE$ equivariant if described on the spatial frame. 
% Because the GUFIC laws, such as force-trajectory tracking and their modification laws, are defined in the end-effector body frame, they are left-invariant. 
% Moreover, since the GUFIC laws, such as force tracking and trajectory modification laws, are all defined on the end-effector body frame, 
% the control law 
This structural property ensures frame-invariant behavior and facilitates seamless integration with $\SE$ visuomotor policies, as shown in \cite{seo2025equicontact}.

\begin{rem}[Further Stability Result]
    From the passivity result, \eqref{eq:time_deriv_storage_passive}, the stability of the system can be proved, similar to \cite{haddadin2024unified}. The proof of stability via Lyapunov's direct method can be conducted using the augmented Lyapunov function with the coupling term as in \cite{seo2023geometric, bullo1999tracking}. 
\end{rem}

\section{Simulation Results}\label{Sec:Simulation}
We implemented the GUFIC control law in the MuJoCo simulation environment
\citep{todorov2012mujoco}, using the Neuromeka Indy7 robot
\citep{tadese2022two}. 
Two scenarios were designed to evaluate force tracking under contact-rich
motion: circular trajectory tracking on a planar surface and line trajectory
tracking on a spherical surface involving full $\SE$ motions. 
We compare GUFIC against geometric impedance control (GIC), which uses the
same geometric impedance structure but does not include the proposed
force-tracking tank augmentation. Since GIC does not explicitly regulate the
contact force, we assume that it only has a rough estimate of the surface
geometry, leading to imperfect force application.

\begin{figure}[t]
    \centering
    \begin{subfigure}{0.49\columnwidth}
        \centering
        \includegraphics[width=1.0\linewidth]{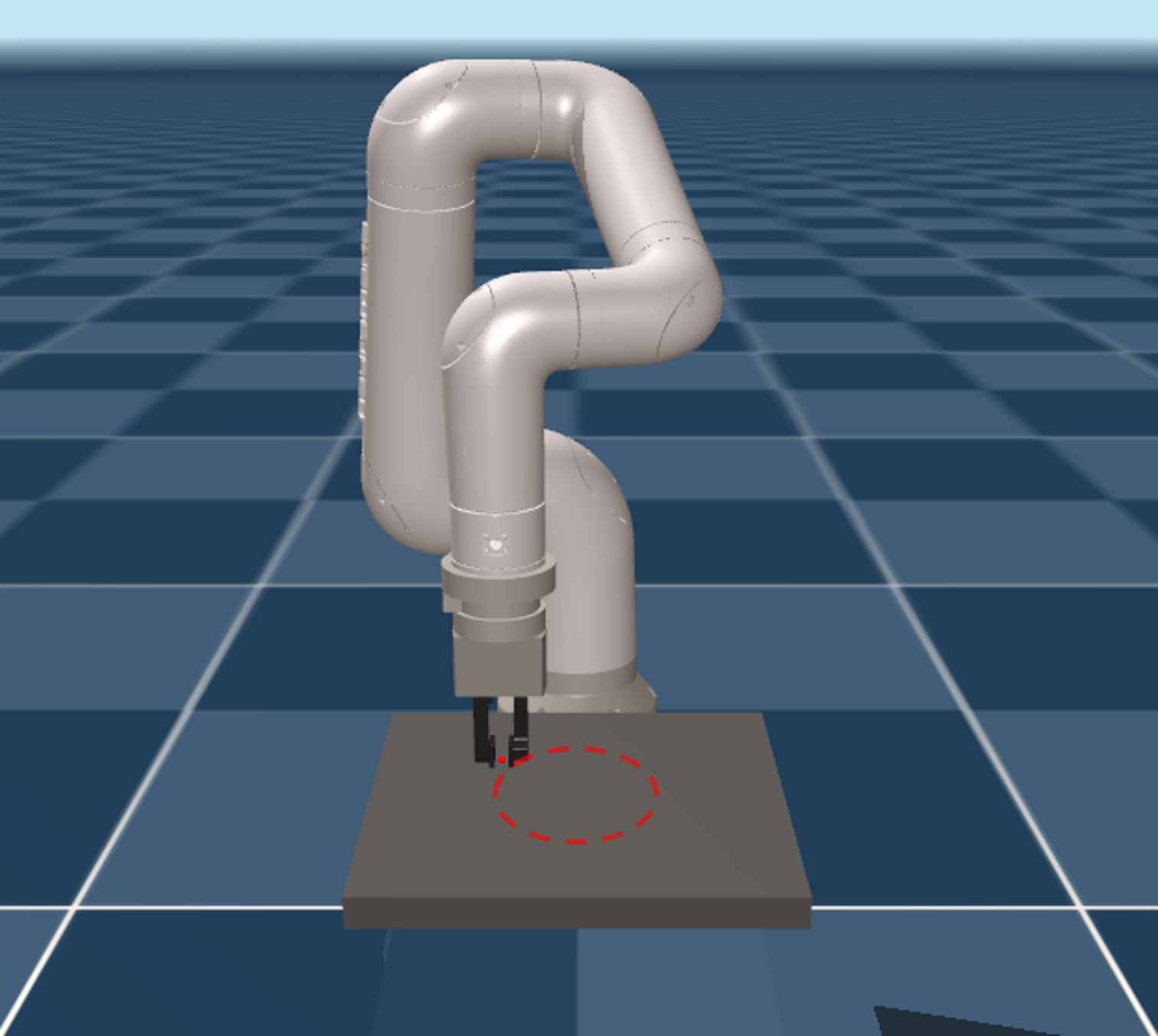}
    \end{subfigure}
    \begin{subfigure}{0.49\columnwidth}
        \centering
        \includegraphics[width=1.0\linewidth]{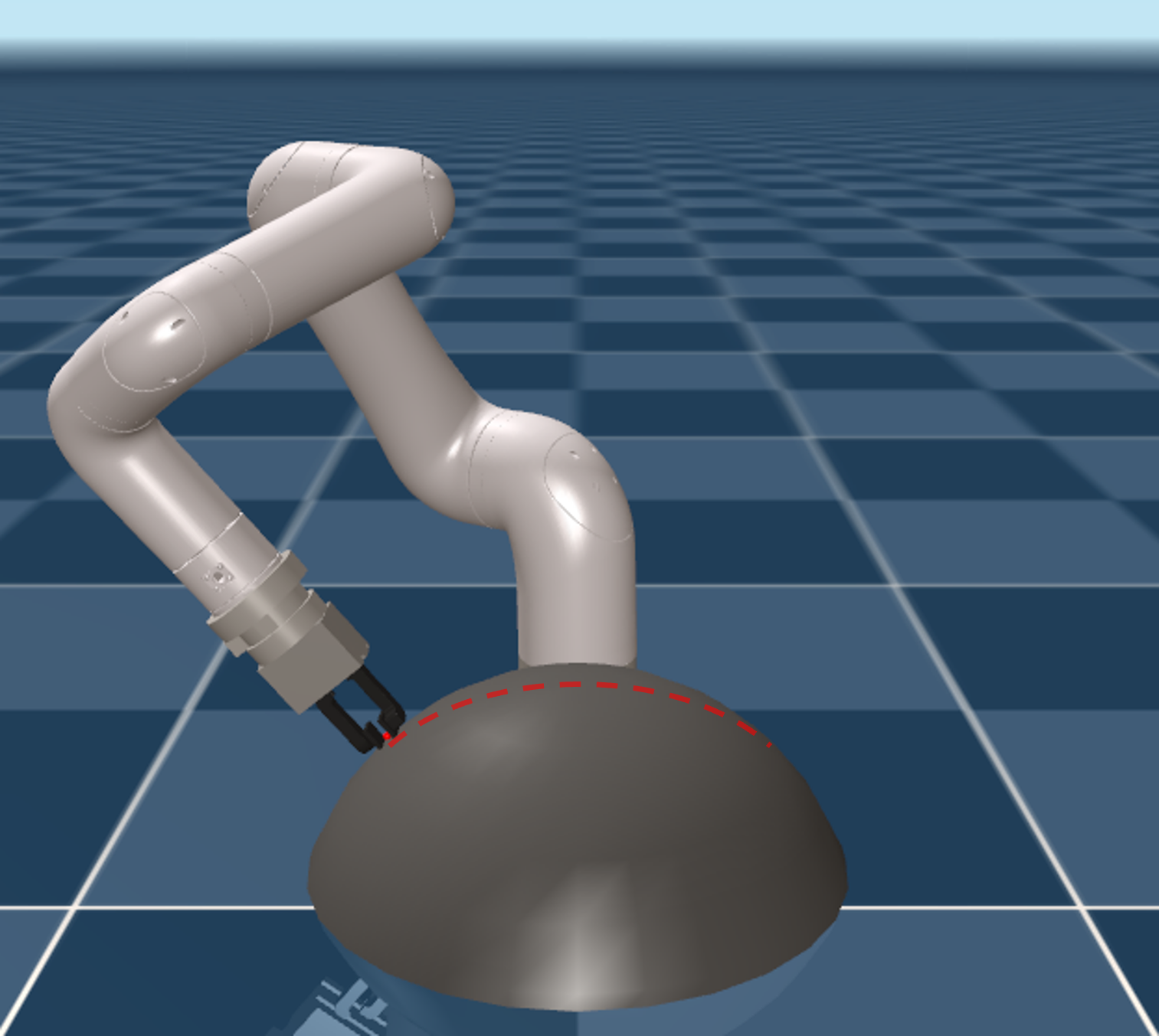}
    \end{subfigure}
    \vspace{-17pt}
    \caption{MuJoCo simulation environments with the Indy7 robot. 
    (Left) Circular tracking on a planar surface. 
    (Right) Line tracking on a spherical surface. 
    Desired trajectories are shown in red dotted lines.}
    \label{fig:mujoco}
    % \vspace{-5pt}
\end{figure}

\subsection{Evaluation Scenarios}

In both scenarios, the robot tracks an $\SE$ reference trajectory while
exerting a desired normal force of $10\,\mathrm{N}$. 
The force-torque sensor was filtered using a $2$\textsuperscript{nd}-order
low-pass filter with a cutoff frequency of $5\,\mathrm{Hz}$, and the
simulation and control loop ran at $1000\,\mathrm{Hz}$. 
The performance metrics are \textbf{translational tracking error}
$\mathrm{RMSE}(p-p_d)$, \textbf{rotational tracking error}
$\mathrm{RMSE}(\tr(I-R_d^TR))$, and \textbf{force tracking error}
$\mathrm{RMSE}(e_f)$.

The quantitative results are summarized in Table~\ref{table:performances}.
GUFIC achieves comparable trajectory tracking to GIC while improving force
tracking performance, especially in the spherical-surface scenario, where inaccurate
surface information causes larger force-tracking error for GIC. The trajectory tracking results are shown in Fig.~\ref{fig:circle_xyz} and Fig.~\ref{fig:sphere_xyz}.
The force tracking results are presented in Figs.~\ref{fig:circle_force} and
\ref{fig:sphere_force_z}. 
For both scenarios, the force and impedance tanks remained positive
throughout execution, confirming that the tank constraints were not violated
during simulation.

\begin{table}[t!]
    \setlength\doublerulesep{0.5pt}
    \renewcommand\tabularxcolumn[1]{m{#1}}
    % \vspace{-3pt}
    \caption{Trajectory and force tracking RMSE.}
    \label{table:performances}
    \vspace{-4pt}
    \centering
    \begin{tabularx}{\linewidth}{
     >{\centering\arraybackslash\hsize=1.0\hsize}X >
     {\centering\arraybackslash \hsize=1.0\hsize}X >
     {\centering\arraybackslash \hsize=1.0\hsize}X >
     {\centering\arraybackslash \hsize=1.0\hsize}X >
     {\centering\arraybackslash \hsize=1.0\hsize}X 
    }
    \toprule[1pt]\midrule[0.3pt]
     Scenarios & Methods & Trans. Err. ($\mathrm{m}$) & Rot. Err & Force Err. ($\mathrm{N}$)\\
    \midrule
    \multirow{2}{*}{Scenario 1} & GUFIC & 0.01989 & 0.01165 & 4.0064 \\
                                & GIC   & 0.01808 & 0.01293 & 4.3606 \\
    \midrule
    \multirow{2}{*}{Scenario 2} & GUFIC & 0.01286 & 0.01156 & 4.0356 \\
                                & GIC   & 0.01547 & 0.01281 & 8.2564 \\
    \midrule[0.3pt]\bottomrule[1pt]
    \end{tabularx}
\end{table}

\begin{figure}[t!]
    % \centering
    \hspace{-8pt}
    \input{Figures/circle_xyz.tex}
    \vspace{-20pt}
    \caption{Trajectory tracking results of GUFIC and GIC for the planar surface scenario.}
    \label{fig:circle_xyz}
    % \vspace{-5pt}
\end{figure}

\begin{figure}[t!]
    \hspace{-8pt}
    \input{Figures/sphere_xyz.tex}
    \vspace{-20pt}
    \caption{Trajectory tracking results of GUFIC and GIC for the spherical surface scenario.}
    \label{fig:sphere_xyz}
    \vspace{-5pt}
\end{figure}

\begin{figure}[t!]
    \hspace{-8pt}
    \input{Figures/circle_force_z.tex}
    \vspace{-20pt}
    \caption{Force tracking result of GUFIC and GIC for the planar surface scenario.}
    \label{fig:circle_force}    
    \vspace{-5pt}
\end{figure}

\begin{figure}[t!]
    \hspace{-8pt}
    \input{Figures/sphere_force_z.tex}
    \vspace{-20pt}
    \caption{Force tracking result of GUFIC and GIC for the spherical surface scenario.}
    \label{fig:sphere_force_z}    
    % \vspace{-5pt}
\end{figure}
\section{Conclusion and Future Works}\label{Sec:Conclusion}
This paper proposes a geometric unified force-impedance control (GUFIC) framework that fully exploits the $\SE$ manifold structure to achieve robust force tracking while ensuring passivity. 
The proposed approach ensures safe interaction with uncertain environments by augmenting energy tanks to the controller. 
Furthermore, the introduction of velocity and force fields resolves the non-causal implementation issues present in earlier frameworks.
The control design inherits $\SE$ invariance and equivariance properties by formulating the unified force-impedance control through differential geometric methods, improving learning sample efficiency. Simulation results have validated the effectiveness of GUFIC in executing complex $\SE$ motions, demonstrating its capability to track trajectories with both position and orientation changes while maintaining the desired force profile. 

Although we assumed in this paper that the velocity and force fields are defined in advance and provided, an interesting question arises as to whether these fields could be learned from experts' demonstrations. In this regard, our future work will focus on learning velocity and force fields via imitation learning using equivariant methods, ensuring that the entire model pipeline guarantees passivity and equivariance. 

{\small
\vspace{-6pt}
\appendix
\section{}
\vspace{-6pt}
\subsection{Proof of Passivity} \label{Sec:proof}
\vspace{-8pt}
Using the control law \eqref{eq:passive_GUFIC}, the modified error
dynamics are
\begin{equation}
    \tilde{M}\dot{e}_V' + \tilde{C}\ev' + K_d \ev'
    + \fg(g,g_d') = F_f' + F_e .
\end{equation}
Let
\begin{equation}
    S = \frac{1}{2}(\ev')^T\tilde{M}\ev' + P(g,g_d') .
\end{equation}
Using \(\dot{P}=\fg(g,g_d')^T\ev'\) and the skew-symmetry of
\(\dot{\tilde{M}}-2\tilde{C}\), we obtain
\begin{align}
    \dot{S}
    &= (\ev')^T \tilde{M}\dot{e}_V'
    + \frac{1}{2}(\ev')^T\dot{\tilde{M}}\ev'
    + (\ev')^T\fg(g,g_d') \nonumber \\
    &= -(\ev')^T K_d \ev' + (\ev')^T F_f' + (\ev')^T F_e .
\end{align}
Since \(\ev'=V^b-V_d'\), this can be rewritten as
\begin{equation}
    \dot{S}
    = -(\ev')^T K_d \ev'
    + (V^b)^T(F_f'+F_e)
    - (V_d')^T(F_f'+F_e).
\end{equation}
From the tank dynamics \eqref{eq:force_tank_state_dyanmics} and
\eqref{eq:impedance_tank_state_dynamics},
\begin{align}
    &\dot{T}_f+\dot{T}_i
    =-\beta_f \gamma_f (V^b)^T F_f
    \!+\! \alpha_f(\gamma_f-1)(V^b)^T F_f \\
    &\!+\!\beta_i\!\left(
    \gamma_i (V_d^*)^T(F_f'\!+\!F_e)
    \!+\!(\ev')^T \! K_d \ev'\right) \!+\!\alpha_i(1\!-\!\gamma_i)(V_d^*)^T \! (F_f'\!+\!F_e) \nonumber.
\end{align}
Using
\begin{equation*}
F_f'=(\gamma_f+\alpha_f(1-\gamma_f))F_f,\qquad
V_d'=(\gamma_i+\alpha_i(1-\gamma_i))V_d^*,
\end{equation*}
the total storage function \(S_{tot}=S+T_f+T_i\) satisfies
% \begin{align}
%     \dot{S}_{tot}
%     =&\underbrace{\gamma_f(1-\beta_f)(V^b)^T F_f}_{\leq 0}
%     +\underbrace{(\beta_i-1)(\ev')^T K_d \ev'}_{\leq 0}
%     \nonumber \\
%     &+\underbrace{\gamma_i(\beta_i-1)(V_d^*)^T(F_f'+F_e)}_{\leq 0}
%     +(V^b)^T F_e . \label{eq:time_deriv_storage}
% \end{align}
\begin{align}
    \dot{S}_{tot}
    &= \gamma_f(1-\beta_f)(V^b)^T F_f
    +(\beta_i-1)(\ev')^T K_d \ev' \nonumber \\
    &\quad +\gamma_i(\beta_i-1)(V_d^*)^T(F_f'+F_e)
    +(V^b)^T F_e \leq (V^b)^T F_e .
    \label{eq:time_deriv_storage}
\end{align}
The first term is non-positive because \(\gamma_f=1\) only when
\((V^b)^T F_f<0\), and otherwise \(\gamma_f=0\). The second term is
non-positive since \(K_d\succ 0\) and \(\beta_i\in\{0,1\}\). The third
term is non-positive because \(\gamma_i=1\) only when
\((V_d^*)^T(F_f'+F_e)>0\), and \(\beta_i-1\leq 0\). Therefore,
\begin{equation}
    \dot{S}_{tot}\leq (V^b)^T F_e ,
\end{equation}
which proves passivity with respect to the port \((V^b,F_e)\).

}
{\footnotesize
\bibliography{bibliography}
}
 
\end{document}